\documentclass{acm_proc_article-sp}

\usepackage{color}

\graphicspath{{Figures/}}

\begin{document}


\title{Evolution of Voronoi based Fuzzy Recurrent Controllers}

\numberofauthors{3}

\author{
\alignauthor Carlos Kavka\\
       \affaddr{Departamento de Informatica}\\
       \affaddr{Universidad Nac. de San Luis}\\
       \affaddr{Ejercito de los Andes 950}\\
       \affaddr{D5700HHW, San Luis, Argentina}\\
       \email{ckavka@unsl.edu.ar}
\alignauthor Marc Schoenauer\\
       \affaddr{\'Equipe TAO, INRIA Futurs}\\
       \affaddr{LRI}\\
       \affaddr{Universit\'{e} de Paris Sud}\\
       \affaddr{91405, Orsay Cedex}\\
       \affaddr{France}\\
       \email{Marc.Schoenauer@inria.fr}
\alignauthor Patricia Roggero\\
       \affaddr{Departamento de Informatica}\\
       \affaddr{Universidad Nac. de San Luis}\\
       \affaddr{Ejercito de los Andes 950}\\
       \affaddr{D5700HHW, San Luis, Argentina}\\
       \email{proggero@unsl.edu.ar}
}

\maketitle

\begin{abstract}
A fuzzy controller is usually designed by formulating the knowledge of a human expert into a set of linguistic variables and fuzzy rules. One of the most successful methods to automate the fuzzy controllers development process are evolutionary algorithms. In this work, we propose the Recurrent Fuzzy Voronoi (RFV) model, a representation for recurrent fuzzy systems. It is an extension of the FV model~\cite{kavka:voronoi} that extends the application domain to include temporal problems. The FV model is a representation for fuzzy controllers based on Voronoi diagrams that can represent fuzzy systems with synergistic rules, fulfilling the $\epsilon$-completeness property and providing a simple way to introduce a priory knowledge. In the proposed representation, the temporal relations are embedded by including internal units that provide feedback by connecting outputs to inputs. These internal units act as memory elements. In the RFV model, the semantic of the internal units can be specified together with the apriori rules. The geometric interpretation of the rules allows the use of geometric genetic operators during the evolution. The representation and the algorithms have been validated in two problems in the area of system identification and evolutionary robotics.
\end{abstract}

\category{}{Genetic Algorithms}{}

\terms{Recurrent Fuzzy Control}

\keywords{Genetic algorithms, Recurrent fuzzy systems, Fuzzy control, Voronoi diagrams, Evolutionary Robotics}

\section{Introduction}

The development of controllers by using fuzzy logic techniques has been subject of theoretical research with many interesting successful applications produced during last years~\cite{babuska:fuzzy}. The main reason is that fuzzy logic controllers (FLC) provide satisfactory performance in face of uncertainty and imprecision~\cite{hagras:type2}, while keeping an equivalence in knowledge representation with other methods like neural networks and automata~\cite{giles:equivalence}. An FLC represents a non linear model as the combination of a set of local linear models, where each one represents the dynamics of a complex system in a single local region~\cite{feng:adaptivecontrol}. Each local model is specified by a fuzzy rule, which defines the local region in which the rule applies through the membership functions used in the antecedent, while the consequent defines the output of the model. Most FLCs can be classified in two categories: the Mamdani type FLC where the output is computed as a combination of fuzzy numbers, and the Takagi-Sugeno (TS) type, where the output is defined as a linear combination of the inputs. In a TS type FLC with $n$ inputs and $m$ outputs, a typical rule has the following form:
\begin{equation}
\begin{array}{lll}
R_i: & \mathbf{if} & x_1 \ \mathbf{is} \ A^1_{i} \ \mathbf{and} \ \ldots
       \mathbf{and} \ x_n \ \mathbf{is} \ A^n_{i} \\
     & \mathbf{then} & y_1 = a_{i0}^1 + \sum_j a_{ij}^1 x_j \\
     & & \ldots \\
     &               & y_m = a_{i0}^m + \sum_j a_{ij}^m x_j \\
\end{array}
\label{eq:tsrule}
\end{equation}
where $x_j (1 \le j \le n)$ are the input variables, $y_j (1 \le j \le m)$ are the output variables, $A^j_i (1 \le j \le n)$ are the fuzzy membership sets and $a_{ij}^k (0 \le j \le n,1 \le k \le m)$ are the real valued parameters that define the linear approximation. The output of the complete FLC is computed by combining the outputs produced by all the rules, weighted by the degree of satisfaction of the antecedents. 

Simple FLCs are usually defined by a trial and error process by using expert knowledge. However, automatic FLC generation methods are preferred for complex control systems. Most FLC structures can be mapped into feed-forward neural networks, allowing the use of neural network learning algorithms to automate the design of FLC based on numerical data as well as on expert knowledge. The combined approach provides advantages from both worlds: the low level learning and computational power of neural networks is joined together with the high level human-like thinking and reasoning of fuzzy systems~\cite{lin:nnfs}. This combination has been very successful and there is a large number of models that combines fuzzy systems with neural networks~\cite{lin:nnfs}, or even with standard PID control~\cite{sun:robotics}~\cite{hojati:hybrid}.

The domain of application of these systems is limited to static problems due to its feed forward network structure~\cite{lee:control}. Most non linear problems in control require the processing of temporal sequences, or in other words, in these problems the output depends on the current input and previous values of inputs and/or outputs. A very interesting approach that considers small order temporal problems with fuzzy logic is proposed in~\cite{carinena:temporal}. However, unless the number of delayed inputs and outputs is known before, it is not possible to define a feed forward model that can process temporal sequences~\cite{juang:tsk}. This is usually the case for most control problems, where this information is not known. However, recurrent structures can deal with this kind of problems. There is a large number of neural network models that have been proposed which are essentially feed forward structures with an extra set of units used to store previous activation values that are connected back to the inputs of other units.

By considering the amount of recurrent neural network models that have been proposed, it is not unexpected to see that most recurrent fuzzy systems are based on neural networks. For example, the model RFNN (Recurrent Fuzzy Neural Network) proposed in~\cite{lee:control} defines recurrent connections in the second layer of the structure, which corresponds to the units that codifies the membership antecedent values. The model RSONFIN (Recurrent Self Organizing Neural Fuzzy Inference Network) proposed in~\cite{juang:inference} perform structure and parameter learning and includes an extra layer of units with recurrent connections that provides a kind of internal memory. The model DFNN (Dynamic Fuzzy Neural Network) proposed in~\cite{mastorocostas:identification} includes recurrent neural networks in the consequent in place of standard linear approximators like in the TS model. The TRFN (Takagi-Sugeno Type Recurrent Fuzzy Network) model proposed in ~\cite{juang:tsk} has an extra unit with recurrent connections for each fuzzy rule, which is responsible of memorizing the temporal history of activations of the rule.  Other models like ~\cite{kasabov:denfis} and~\cite{zhang:nonlinear} follow similar approaches. In most cases, both supervised learning and non gradient based algorithms, like genetic algorithms or reinforcement learning, have been used to build or enhance the models.

Even if these recurrent models are successful in supporting learning of temporal sequences, in most cases the logic interpretation of recurrent units is not considered. In this work, we propose a recurrent structure for fuzzy systems based in the Fuzzy Voronoi (FV) method proposed in~\cite{kavka:voronoi}, which allows the definition of recurrent fuzzy systems with a clear interpretation of recurrent units. The proposed Recurrent Fuzzy Voronoi (RFV) structure consists in a set of rules, where the antecedent of the rules are determined by multidimensional membership functions defined in terms of Voronoi regions. The RFV model includes external and internal variables with recurrent connections that allow the processing of temporal sequences of arbitrary length. Genetic algorithms are proposed as a design tool, since they do not require derivative information, which in most control problems is unavailable or costly to obtain.

This paper is organized as follows. Section~\ref{sec:structure} describes the structure of the RFV model and the details on the geometrical basic structure. In section~\ref{sec:algorithm}, the design of RFV models with genetic algorithms is analyzed. The properties of the proposed representation are discussed in section~\ref{sec:properties}. In section~\ref{sec:experiments}, two experiments with the RFV model are detailed. Finally, conclusions are presented in section~\ref{sec:conclusions}.

\section{The RFV model}
\label{sec:structure}

In this section, the structure of the RFV model is presented, the fuzzy reasoning strategy is explained and the details on the computation of the membership functions are introduced.

\subsection{Structure}

\begin{figure*}
\centering
\includegraphics[scale=0.5]{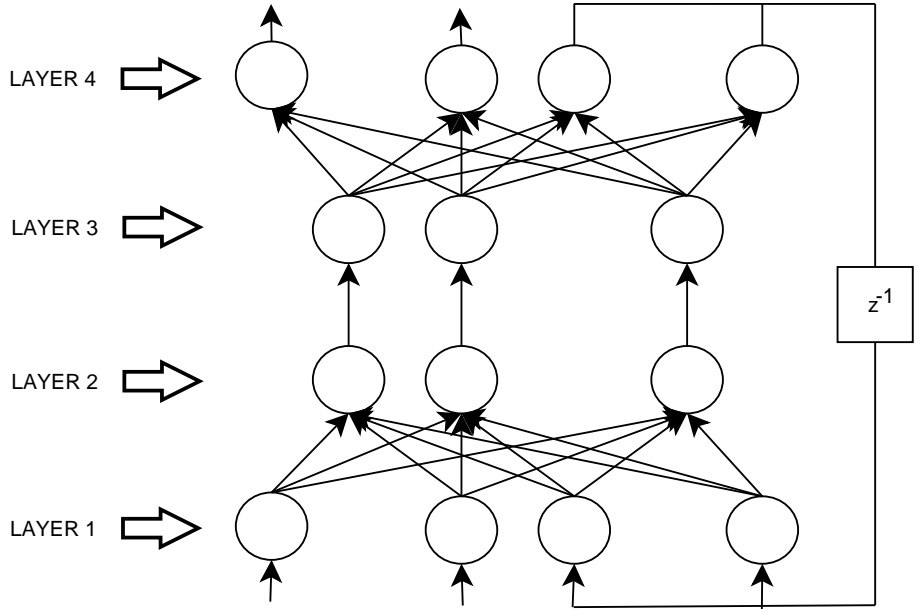}
\caption{The structure of the RFV model}
\label{fig:model}
\end{figure*}

A schematic diagram of the model is shown in figure~\ref{fig:model}, which is organized in four layers and consists of $l$ input variables, $r$ internal variables, $m$ output variables and $\omega$ rules. 
Units in layer 1 are called input units. There are two types of input units: external inputs and internal units that are used also as standard inputs in rule definition. Units in layer 2 are called partition units. They act as multidimensional fuzzy membership functions. Units in layer 3 are called rule units. Each fuzzy rule in the fuzzy system has a corresponding rule unit. There is a one to one correspondence with units in layer 2. Units in layer 4 are called output units. They compute the outputs as a weighted linear combination of input units, generating both the external outputs and the values of the internal units to be made available as inputs in the next time step.

The function of each type of unit is described below. In the descriptions, the external input vector of size $l$ is denoted by $x$, the internal vector of size $r$ is denoted by $h$, the output vector of size $m$ is denoted by $y$, the complete input vector for the rules of size $l+r$ is denoted by $I = x:h$ and the complete output vector produced by the rules of size $m+r$ is denoted by  $O = v:y$, where $:$ identifies the concatenation operator.

\begin{description}
\item[Layer 1]: No computation is performed in this layer. External input values $x$ and previous values of internal units $y$ are transmitted to the units in layer 2.
\item[Layer 2]: The $k$-th unit in this layer computes the fuzzy membership value $\mu_{S_k}(I)$ of the input vector $I$ to the multidimensional fuzzy set $S_k$ associated to the $k$-th rule.  More details on this computations are provided in section~\ref{sec:fuzzyvoronoi}.
\item[Layer 3]: Units in this layer compute a linear combination of input values based on the parameters specified by each rule, weighted by the corresponding degree of activation, as usual in TS fuzzy systems. Note that this units produce $m + r$ outputs. The output produced by the unit $k$ that corresponds to the output variable $i$ is:
\begin{equation}
O_i^k = ( a_{k0}^i + \sum_j a_{kj}^i I_j ) \mu_k(I)\\
\label{eq:layer3}
\end{equation}
where the $a_{kj}^i$ are the real valued parameters that compute the linear combination of input values associated to the rule $k$ for the output variable $i$.
\item[Layer 4]: Units in this layer compute the output vector $O$ by computing the summation of the corresponding outputs produced by each rule. That is:
\begin{equation}
O_i = \sum_k O_i^k\\ 
\end{equation}
\end{description}

\subsection{Fuzzy reasoning model}

The RFV model performs fuzzy inference by using rules defined as follows:
\begin{equation}
\begin{array}{lll}
R_k: & \mathbf{if} & I \ \mathbf{is} \ A_{k} \\
     & \mathbf{then} & O_1 = a_{k0}^1 + \sum_j a_{kj}^1 I_j \\
     &               & \ldots \\
     &               & O_{m+r} = a_{k0}^{m+r} + \sum_j a_{kj}^{m+r} I_j \\
\end{array}
\end{equation}
It can be noted that there is a strong similarity with the standard definition of TS rules given in equation~\ref{eq:tsrule}. Except for the fact that now a single multidimensional set is used for membership, the main difference is that the fuzzy inference involves $r$ terms in the input vector  that are output values produced in the previous time step. The fuzzy system is a dynamic fuzzy inference system with the inferred values $v_i$ produced in time $t+1$ given by:
\begin{equation}
v_i(t+1) = \sum_k O_i^k(t)
\end{equation}
where the computation of the value $O_i^k(t)$ (see equation~\ref{eq:layer3}) involves the input vector $x(t)$ at time $t$ and the internal values $y(t-1)$ defined in time $t-1$.

\subsection{Membership computation}
\label{sec:fuzzyvoronoi}

The domain partition strategy is based on Voronoi diagrams.  A Voronoi diagram induces a subdivision of the space based on a set of points called {\em sites}. Formally~\cite{berg:compgeometry}, a Voronoi diagram of a set of $p$ points ${\cal P} =\{P_1, \ldots, P_p\}$ is the subdivision of the plane into
$p$ cells, one for each site in ${\cal P}$, with the property that a point $M$ lies in the cell corresponding to a site $P_i$ if and only if the distance between $M$ and $P_i$ is smaller than the distance between $P$ and all other $P_j$ ($j \ne i$).
\begin{figure}
\centering
\begin{tabular}{cc}
\includegraphics[scale=0.4]{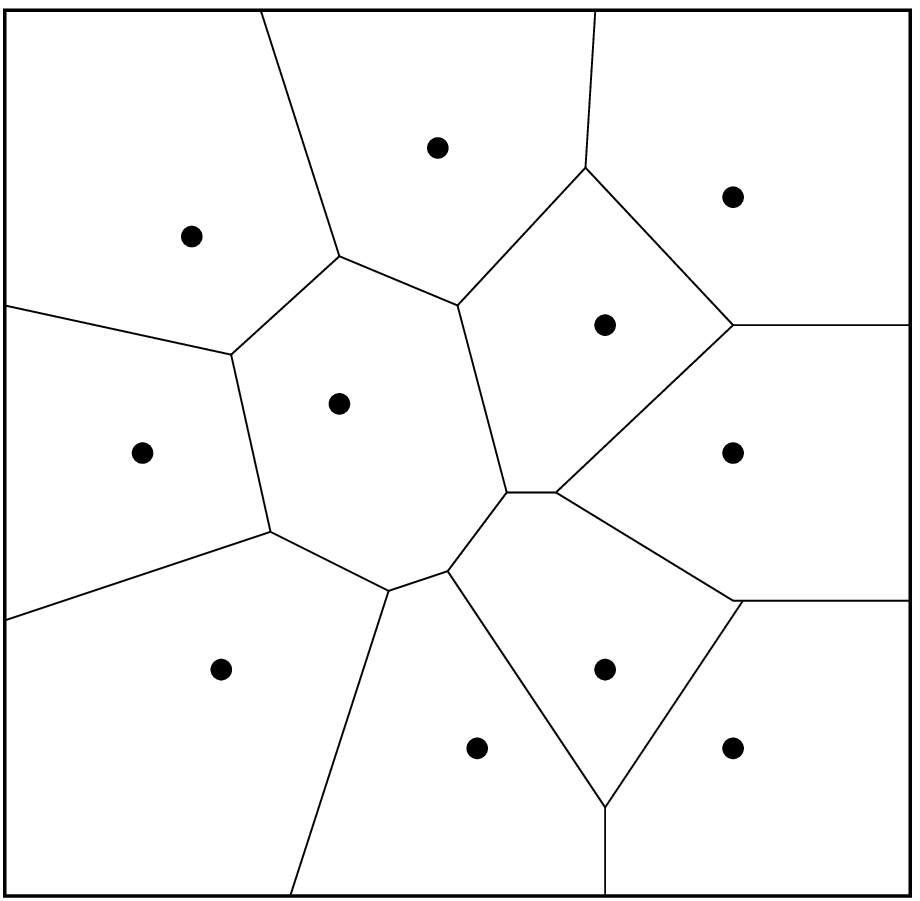} &
\includegraphics[scale=0.4]{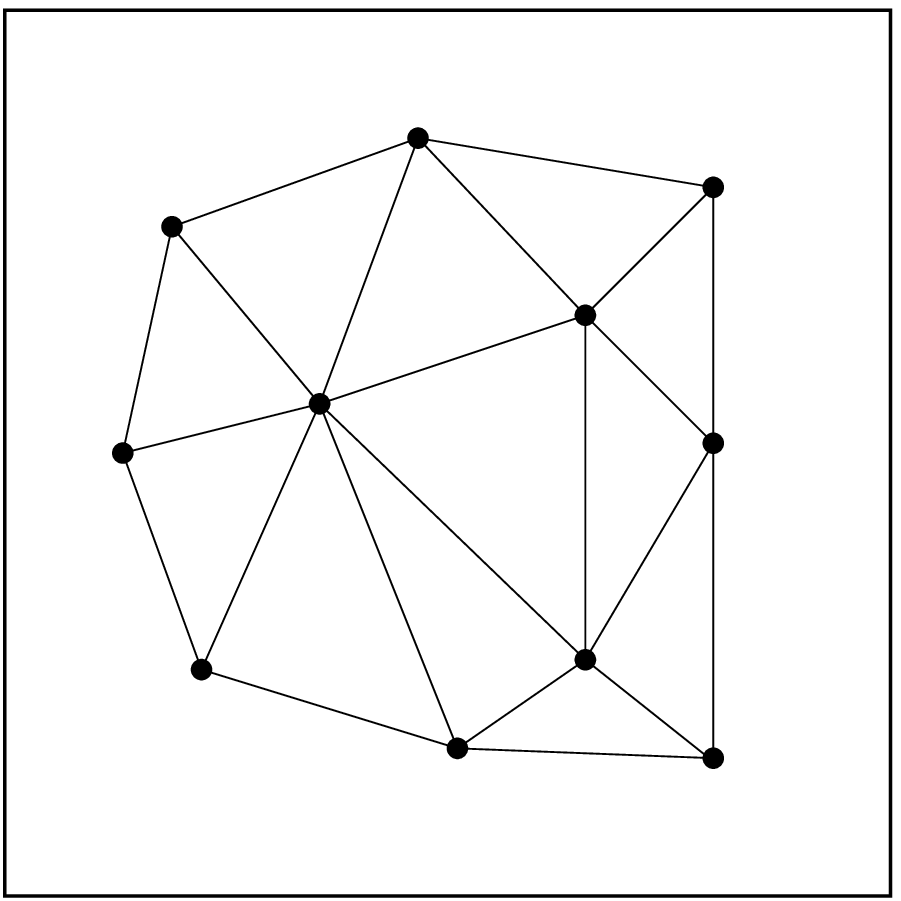} \\
\end{tabular}
\caption{An example of a Voronoi diagram (left) and the corresponding Delaunay triangulation (right) for a set of points in $R^2$}
\label{fig:voronoiexample}
\end{figure}
A related concept is the so called Delaunay triangulation $\mathcal{T}$, defined as the maximal planar subdivision (i.e. a subdivision such that no edge connecting two vertexes can be added to $S$ without destroying its planarity) whose vertex set is $\cal P$ and such that the circumcircle of any triangle in $T$ does not contain any point of $\cal P$ in its interior. Figure~\ref{fig:voronoiexample} illustrates an example of a Voronoi diagram and its corresponding Delaunay triangulation in $R^2$. Note that these definitions can be straightforwardly extended to $R^n$, with $n \geq 2$ -- all details can be found in \cite{berg:compgeometry}.

\begin{figure}
\centering
\begin{tabular}{cc}
\includegraphics[scale=0.7]{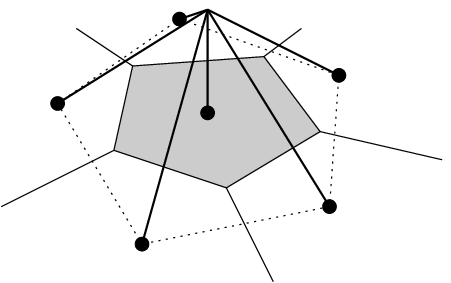} &
\includegraphics[scale=0.7]{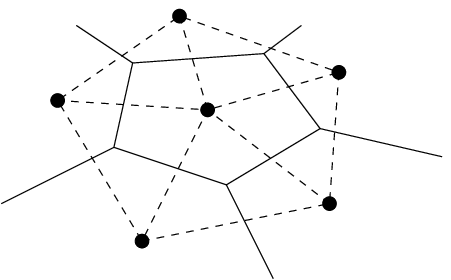} \\
(a) & (b) \\
\multicolumn{2}{c}{\includegraphics[scale=0.9]{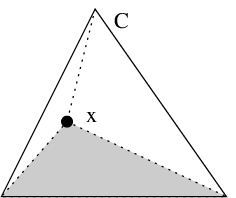}} \\
\multicolumn{2}{c}{(c)} \\
\end{tabular}
\caption{An example of a (a) joint fuzzy set for a single Voronoi region for $n = 2$, where the membership value is represented in the z-axis, and a (b) Voronoi diagram (solid line) and its corresponding Delaunay triangulation (dotted line) for $n = 2$. The graphic (c) shows an example of the membership computation for $n = 2$. The outer triangle corresponds to the simplex defined by the Delaunay triangulation to which $x$ belongs. The membership value corresponds to the area of the shadowed triangle. Note that the value of the area is 1 when $x$ is equal to $C$ and it goes down linearly to 0 on the side of the triangle opposite to $C$}
\label{fig:jointfuzzyset}
\end{figure}

The FV representation~\cite{kavka:voronoi} considers joint fuzzy sets defined from a Voronoi diagram ${\cal P} =\{P_1, \ldots, P_p\}$. There are as many rules as Voronoi sites. The fuzzy set $S_k$ is defined as by its multivariate membership function $\mu_k$ that takes its maximum value 1 at site $P_k$, and decreases linearly to reach value 0 at the centers of all neighbor Voronoi sites. An example of such a joint fuzzy set is shown in figure~\ref{fig:jointfuzzyset}-a for $n = 2$.

Formally, the membership value of the input vector $I$ to the joint fuzzy set $S_k$ is defined by:
\begin{equation}
\mu_{S_k}(I) =
  \left\{
    \begin{array}{ll}
       l_C(I) & x \in V_k \\
      0 & \textrm{elsewhere}. \\
    \end{array}
  \right.
\label{membershipEquation}
\end{equation}
where $C=P_k$ is the Voronoi site defining $S_k$ and the Voronoi cell $V_k$, and $l_C(I)$ is the barycentric coordinate of $I$ in the simplex $T_C(I)$ of the Delaunay triangulation of $\cal P$ that has $C$ as a vertex and to which $I$ belongs. Figure~\ref{fig:jointfuzzyset}-b shows an example of the Voronoi diagram and the associated Delaunay triangulation. On Figure~\ref{fig:jointfuzzyset}-c,
the barycentric coordinate $l_C(I)$ corresponds to the (normalized) gray area (volume if $n > 2$) of the sub-simplex formed by $I$ and vertexes of  simplex $T_C(I)$ but $C$. Note that a very large triangle containing all points in the domain is defined in such a way that there are no open Voronoi regions in the input domain.\\

\section{RFV design with evolutionary algorithms}
\label{sec:algorithm}

Evolutionary algorithms are selected as the optimization tool for RFV controller design, since they have been very successful on problems where training data or gradient information is very difficult or costly to obtain, like most control problems. A floating point coding scheme is selected, where each individual (or chromosome) represents all free parameters of the RFV controller as a variable length vector of floating point values. An individual $I$ with $\omega$ rules is defined as the vector:
\begin{equation}
\mathit{Ind} =  R_1:\ldots:R_\omega 
\end{equation}
where each sub-vector $R_i (1 \le i \le \omega)$ is defined as the floating point vector:
\begin{equation}
R_i = P_i^{1}:\ldots:P_i^{l+r}:a^1_{i0}:\ldots:a^{m+r}_{i(l+r)} 
\end{equation}
where the $P_i^j$ are the coordinates of the site and $a_{kj}^i$ are the real valued parameters associated to the rule $R_i$. The evolutionary algorithm is described in details in~\cite{schoenauer:voronoi,kavka:funcapp}. The
crossover operator is based on geometrical exchange of Voronoi sites between both parents with respect to a random hyperplane. The mutation operator can either modify the parameters of a particular rule by some standard Gaussian mutation, or add or delete a Voronoi site, i.e. a rule (see section \ref{sec:properties}).  Practical details on the algorithms, including all parameters,  will be given in section \ref{sec:experiments}. But before experimentally validating the FV representation, next section will discuss some of its properties.

\section{Properties}
\label{sec:properties}

First of all, the RFV representation belongs to the class of approximative representations~\cite{babuska:fuzzy}, where each fuzzy rule defines its own fuzzy sets. It also provides continuous output, as most fuzzy systems. However, it also has a number of useful properties, that we shall now discuss in turn. 

\vspace{1em}
\textbf{$\epsilon$-completeness property}: All RFV-based fuzzy systems defined with the RFV representation fulfills the completeness property at any required level, which establishes that any input must belong to at least one fuzzy set with a membership value not smaller than a threshold value $\epsilon$:
\begin{equation}
\forall x \in U \: \exists A \in \{ A_1,\ldots,A_n \} \: \mu_A(x) \geqslant \epsilon.
\end{equation}
For the RFV representation, it is clear from the definition of the membership function of equation (\ref{membershipEquation}) that this property will hold with $\epsilon = \frac{1}{2}$, as $l_C(x)$ will be above 0.5 if $x$ lies in the Voronoi cell defined by $C$.

This property guarantees an adequate representation for every input point, since there is always a rule that is applied with at least a known value of membership. 

\vspace{1em}
\textbf{No need for genetic repair algorithms}: Since it is not possible to define wrong or non complete fuzzy systems, the fuzzy systems produced by applying mutation or crossover operators are always valid control systems.

\begin{figure}  
\centering
\begin{tabular}{cc}
\includegraphics[scale=0.18]{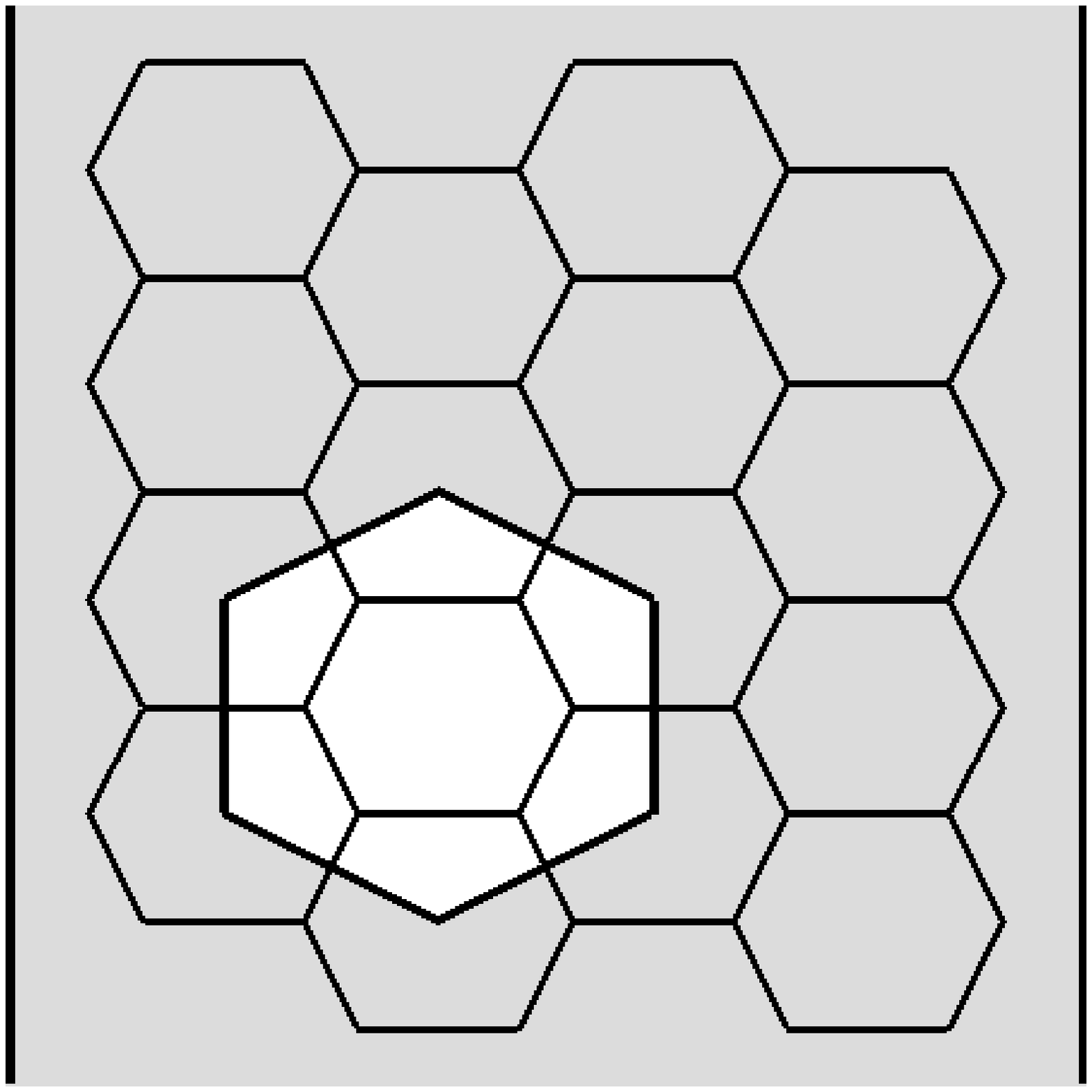} &
\includegraphics[scale=0.18]{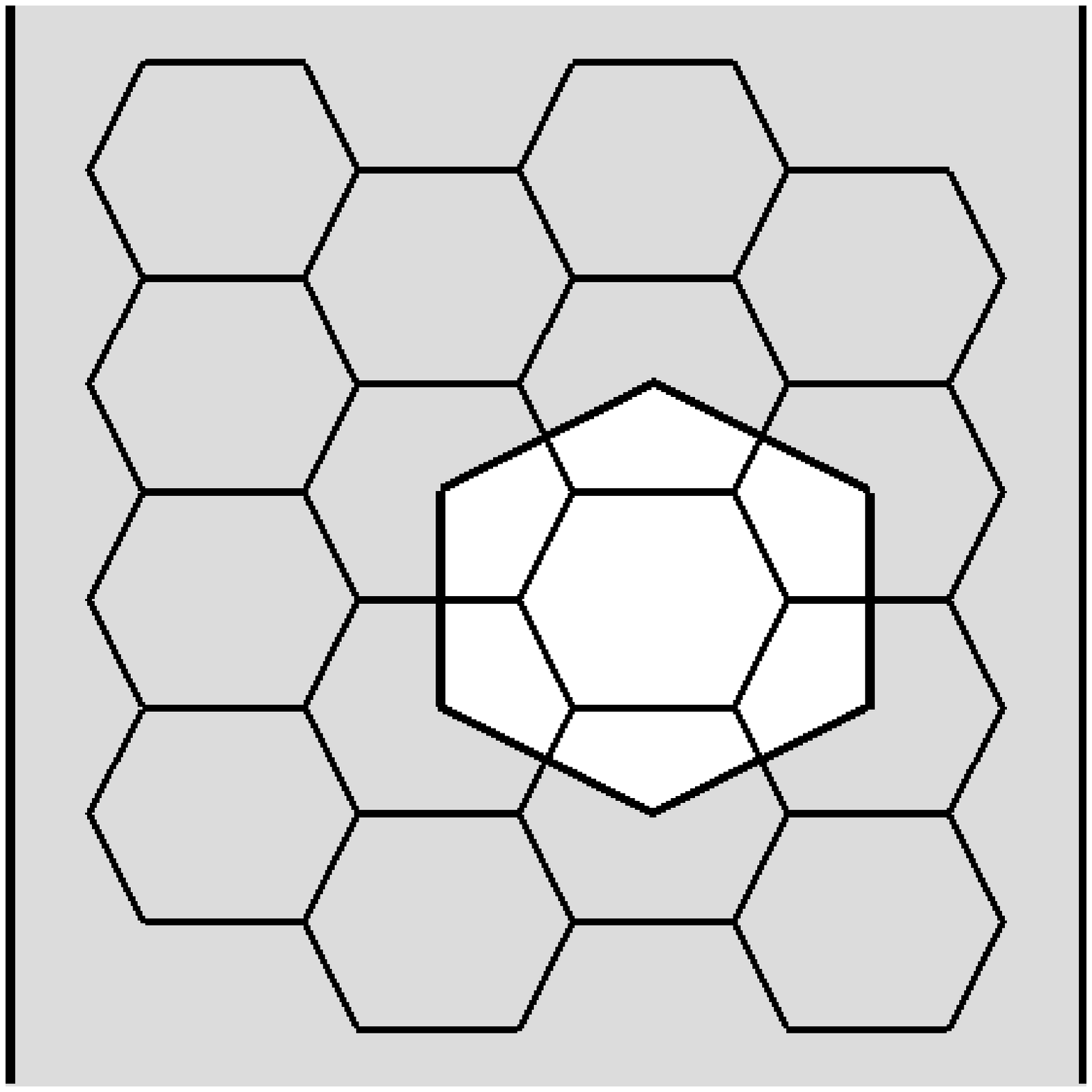} \\
(a) & (b) \\
\includegraphics[scale=0.18]{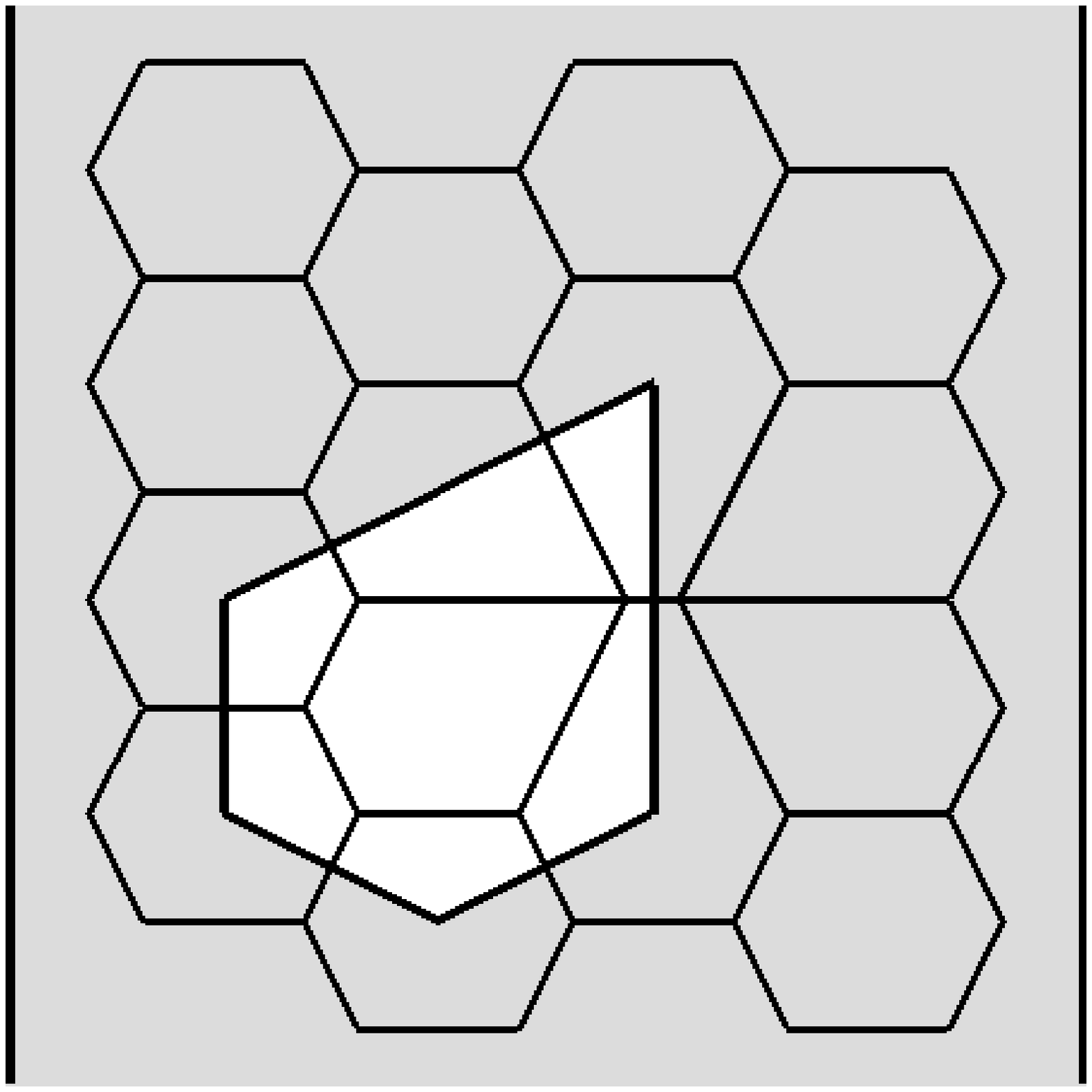} &
\includegraphics[scale=0.18]{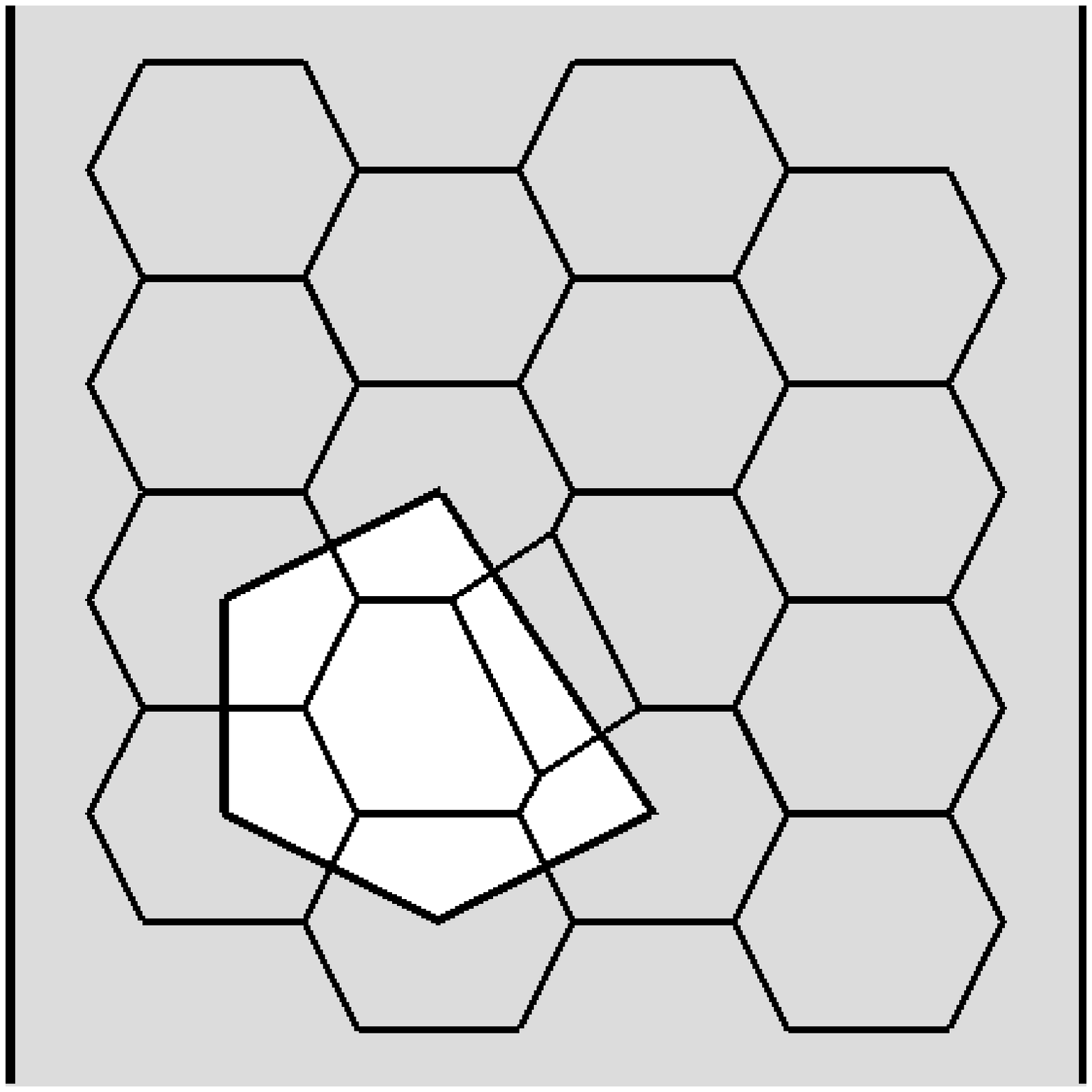} \\
(c) & (d) \\
\end{tabular}
\caption{ Diagram (a) shows the application area of a fuzzy rule. Diagram (b) shows the application area of one of its neighbor rule. Diagram (c) shows the application area of the rule of diagram (a) when the rule of diagram (b) is removed, and diagram (d) the application area of the rule of diagram (a) when a rule is added between both rules.} 
\label{fig:area}
\end{figure}

\vspace{1em}
\textbf{Adaptive fuzzy rules}: The influence on the output of a particular fuzzy rule in the RFV representation does not only depend on the rule itself, it also depends on all neighbor rules. The area of application $\mathcal{A}_k$ of a fuzzy rule $R_k$ is defined as the union of all Delaunay regions which contain the point $P_k$, center of the rule $R_k$. Formally:
\begin{equation}
\mathcal{A}(R_k) = \bigcup_{P_k \in D_j} D_j \: \: \: \: \: D_j \in
D = \{ D_1,\ldots,D_{\gamma}\}.
\end{equation} 
where $P_k$ is the center of the rule $R_k$ and $D = \{ D_1,\ldots,D_{\gamma}\}$ is the Delaunay partition of the set ${\cal P} = \{P_1,\ldots,P_{p}\}$. Figure~\ref{fig:area} shows an example of the
application area of some rules in a regular partition, and illustrates the interdependency of application areas of neighboring rules when some rules are removed or added.

The evolutionary algorithm evolves individuals that represent complete fuzzy systems defined by a set of fuzzy rules that are synergistically related, and not fuzzy systems defined with a set of independent fuzzy rules. The variation operators hence modify the application areas of all fuzzy rules, while still maintaining the required $\epsilon$-completeness level.

\vspace{1em}
\textbf{Adaptive apriori rules}:
In most fuzzy systems, the user can incorporate apriori knowledge by manually defining fuzzy sets and the corresponding fuzzy rules. This process implies that some restriction on the output values and the partition of the input space is introduced in the evolutionary process, but the expected benefit is that the evolutionary process, biased toward hopefully good parts of its search space, will converge faster to better solutions.

Similarly, the RFV representation allows the definition of apriori rules, i.e. fixed Voronoi sites that will not be modified by evolution. But one big advantages of the RFV representation is that the expert does not need to specify the application area of such rules: thanks to the synergistic effect described above, the evolutionary process, by adding rules more or less close to the apriori rules will also tune its domain of application -- as will be clear on the experimental results in section~\ref{sec:experiments}. 

\vspace{1em}
\textbf{Recurrent rules}:
The rules defined in the RFV controller are standard TS type fuzzy rules, with their own inputs and outputs. The complete system is recurrent because some outputs are connected to inputs, but each rule by itself is a standard TS type fuzzy rule. This fact contributes to provide a clear interpretation of the rules and make easy to define the apriori rules for the RFV controller. This approach contrasts with other models like RFNN~\cite{lee:control}, RSONFIN~\cite{juang:inference}, DFNN~\cite{mastorocostas:identification} or the TRFN~\cite{juang:tsk}, where the rules themselves include backward connections. The recurrent connection model is similar to the NFSLS approach proposed in~\cite{mouzouris:nfsls}, except that only a single output is considered and standard fuzzy partition is performed in the input domain. Section~\ref{sec:experiments} will introduce examples that show that this way of defining recurrent rules allows easy introduction of apriori knowledge.

\section{Experiments}
\label{sec:experiments}

In this section, the evolutionary approach to design RFV systems is evaluated in two problems. The first one is a system identification problem, where the outputs of the system are function of past inputs and outputs. This problem is introduced in order to compare the approach with other methods. The second problem is more interesting: it is an evolutionary robotic problem~\cite{nolfi:robotics}, where the ability to introduce apriori knowledge in the form of recursive rules is demonstrated.

\subsection{System identification}

\begin{figure}  
\centering
\includegraphics[scale=0.6]{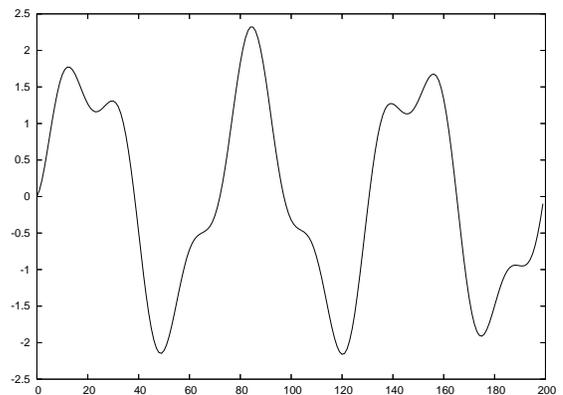}
\caption{The control signal in the system identification problem.} 
\label{fig:exp1}
\end{figure}

The controlled plant is the same as used in the example 3 in~\cite{juang:tsk} and is given by:
\begin{equation}
y_p(t+1) = \frac{y_p(t)y_p(t-1)(y_p(t)+2.5)}{1+y^2_p(t)+y^2_p(t-1)} + u(k)
\end{equation}
where $y_p(t)$ and $u(t)$ are respectively the output and the input at time $t$. The desired output is defined by 250 pieces of data obtained from:
\begin{equation}
\begin{array}{rl}
y_r(t+1) = & 0.6 y_r(t) + 0.2 y_r(t-1) + 0.2 sin(2 \pi t / 25) \\
           &+ 0.4 sin(\pi k / 32) \\
\end{array}
\end{equation}
The  test signal used for evaluation is shown in figure~\ref{fig:exp1}.

In the experiments, the population size is set to 50, the probability of Voronoi crossover is set to 0.8, Voronoi mutation to 0.3, mutation for addition and removal of Voronoi sites to 0.1, selection is performed by tournament, elitism is used and the number of generations is set to 1200.  The fitness is defined as the RMS error. The results for the best and average RMS error over 50 runs are listed in table~\ref{table:exp1}. The table shows also the results obtained with the TRFN and RFNN models as presented in~\cite{juang:tsk}. The comparison has to be considered with extreme care since, even if a careful selection of parameters was performed to replicate the experiments, there are some important differences in the models. For example, the RFV model uses variable length individuals, while the other methods use fixed length individuals. The main implication is that the number of fuzzy rules in the RFV model is determined by evolution, while in the other models has to be defined in advance. The number of rules in the other experiments is set to 4, which also determine the number of hidden units, since there is a one to one correspondence between hidden units and fuzzy rules. In the RFV model, the number of hidden units is independent and it is set to 1. Results in CPU time are not provided (even if they are provided in~\cite{juang:tsk}) since there is no information on the hardware used to run the experiments and the comparisons will not be fair.
\begin{table}
\centering
\caption{RMS error for the system identification experiment with one internal unit}
\begin{tabular}{|c|c|c|c|c|c|} \hline
\multicolumn{2}{|c|}{RFNN+GA} & \multicolumn{2}{c|}{TRFN+GA} & \multicolumn{2}{c|}{RFV} \\ \hline
mean   & best   & mean   & best   & mean   & best   \\ \hline
0.3911 & 0.0850 & 0.0910 & 0.0536 & 0.0775 & 0.0235 \\ \hline
\end{tabular}
\label{table:exp1}
\end{table}

\subsection{Evolutionary robotics}

A problem defined in the area of evolutionary robotics~\cite{nolfi:robotics} has been selected to validate the RFV model. As a test base for experiments, a simulated Khepera robot~\cite{michel:khepera} was used for experimentation. A Khepera robot has 8 sensors that can be used to measure proximity of objects and ambient light levels, and two independent motors to control the speed and direction of the robot. The problem consists in drive the robot avoiding collisions, starting from a fixed initial position, to a target position that depends on light based signals that are set to on or off status in the trajectory. The presence of an illuminated signal (on status) indicates to the robot that it has to turn left in the next intersection, and its absence (or off status) that it has to turn right. The controller needs internal memory, since the light signal is not present in the intersection, but in a previous (and may be distant) point in the trajectory. The controller has to learn also to \emph{forget} light signals that affected the behavior in previous intersections and have not to be considered in other point of the trajectory.

The fitness of a RFV controller is computed in a similar way as in~\cite{nolfi:robotics}, evaluating the controller in \emph{e} different scenarios. Each scenery defines initial and target positions, and include path intersections where light signals determine the expected trajectory of the robot. The fitness is accumulated at every step of the robot proportionally to the speed, inversely proportional to the distance to the target point and reduced when the robot travels near obstacles, in order to favor navigation without collisions. The fitness accumulation is stopped when the robot bumps an obstacle, or it reaches a maximum number of steps \emph{s}. The total fitness is the average of the values obtained in the \emph{e} scenarios. Formally, the fitness is defined as follows:
\begin{equation}
\textrm{fitness}(I) = \frac{1}{e s} \sum_{i=1}^{e} \sum_{t=1}^{s} v(t) * (1 - a(t)) * (1 - d(t))
\end{equation}
where $t$ is the time step, $v(t)$ is the normalized forward speed (summation of the speed of both motors), $a(t)$ is the normalized maximum activation of the sensors~\cite{nolfi:robotics} (for example, $a(t)=1$ implies a collision) and $d(t)$ is the normalized distance to the destination point (for example, $d(t)=0$ implies that the target has been reached) . This function assigns larger values to individuals that travel at the highest speed, in a trajectory that follows (when possible) a straight line, as far as possible to obstacles and minimizing the distance to the target point. 

The controllers are defined with five inputs, two outputs and one internal variable. The inputs are, respectively, the average of the two left sensors, the two front sensors, the two right sensors, the two back sensors and an average of ambient light as measured by all sensors. The outputs correspond to the speed of the two motors. Note that the presence of an internal variable forces the rules to be defined with six inputs and three outputs (see figure~\ref{fig:model}). In the experiments, the population size is set to 50, the probability of Voronoi crossover is set to 0.8, Voronoi mutation to 0.3, mutation for addition and removal of Voronoi sites to 0.1, selection is performed by tournament, elitism is used and the number of generations is set to 200. The performance of the individuals is measured in $e=4$ scenarios with three intersections and all combinations of light signals, evaluated in at most $s=500$ time steps.
\begin{figure}
\centering
\begin{tabular}{c}
\includegraphics[scale=0.3]{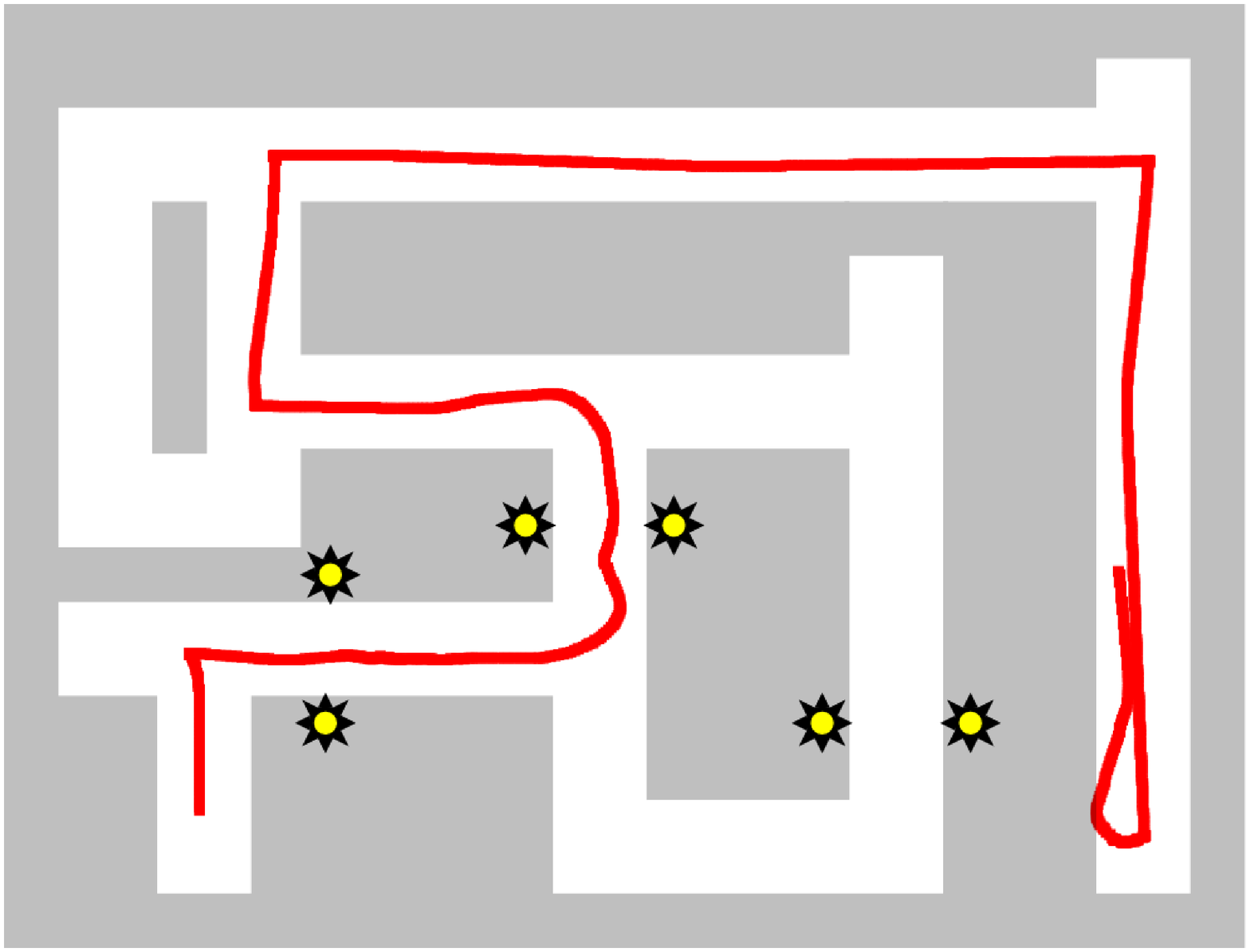} \\
(a) \\
\includegraphics[scale=0.3]{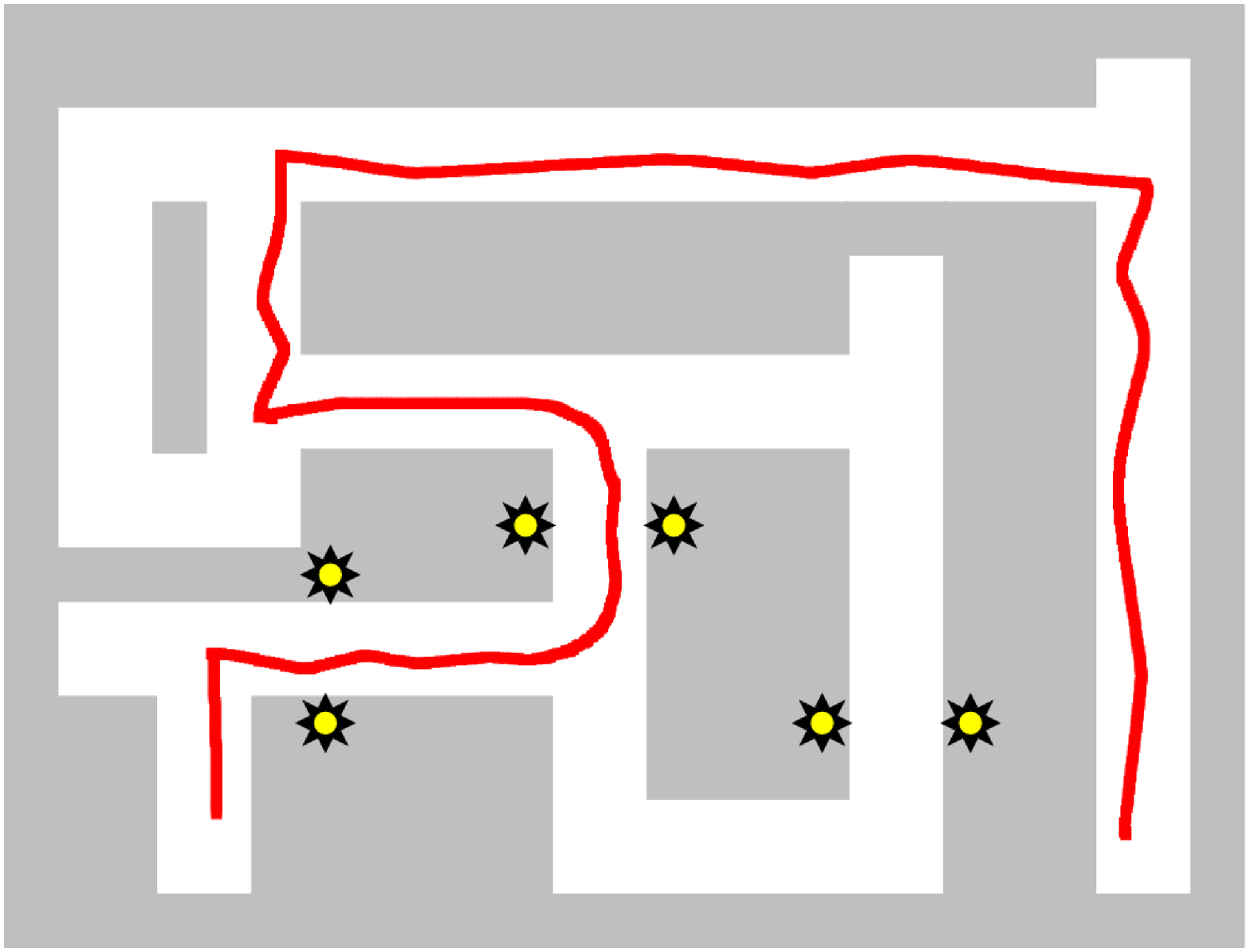} \\
(b) \\
\end{tabular}
\caption{The performance of the best controller (a) without apriori knowledge and (b) with apriori knowledge when evaluated in a scenery not used during evolution.}
\label{fig:evaluation}
\end{figure}

The experiments were performed also by using apriori knowledge. The rules defined beforehand and inserted as explained in section~\ref{sec:properties} are shown in table~\ref{table:rules}.
\begin{table*}
\centering
\begin{tabular}{c|c|c|c|c} \hline
\emph{Rule} & \emph{site} & $v_1$ & $v_2$ & $y_1$ \\ \hline
& {\emph{L},\emph{C},\emph{R},\emph{B},\emph{G},$y_1$} &
{\bf $a^1_0$,$a^1_1$,$a^1_2$,$a^1_3$,$a^1_4$,$a^1_5$,$a^1_6$} &
{\bf $a^2_0$,$a^2_1$,$a^2_2$,$a^2_3$,$a^2_4$,$a^2_5$,$a^2_6$} &
{\bf $a^3_0$,$a^3_1$,$a^3_2$,$a^3_3$,$a^3_4$,$a^3_5$,$a^3_6$} \\ \hline
$R_1$ & 0,0,0,0,0,0 & 1,0,0,0,0,0,0 & 1,0,0,0,0,0,0 & 0,0,0,0,0,0,0 \\
$R_2$ & 0,0,0,0,0,1 & 1,0,0,0,0,0,0 & 1,0,0,0,0,0,0 & 1,0,0,0,0,0,0 \\
$R_3$ & 0,0,0,0,1,0 & 1,0,0,0,0,0,0 & 1,0,0,0,0,0,0 & 1,0,0,0,0,0,0 \\
$R_4$ & 0,0,0,0,1,1 & 1,0,0,0,0,0,0 & 1,0,0,0,0,0,0 & 1,0,0,0,0,0,0 \\
$R_5$ & 0,1,0,0,0,0 & 1,0,0,0,0,0,0 & 0,0,0,0,0,0,0 & 0,0,0,0,0,0,0 \\
$R_6$ & 0,1,0,0,0,1 & 0,0,0,0,0,0,0 & 1,0,0,0,0,0,0 & 0,0,0,0,0,0,0 \\ \hline
\end{tabular}
\caption{Apriori rules. The value of the site corresponds to the center of the Voronoi region defined by the rule. It is specified by the normalized values of the left (L), center (C), right (R), back (B) and light (G) sensors, and the internal variable $y_1$. The values $a^i_j$ correspond to the parameters used to define the approximators.}
\label{table:rules}
\end{table*}
The semantic of the rules is defined by considering the internal variable $y_1$ as a flag that indicates if a light signal was seen before. The first four rules correspond to the situation where there are no obstacles near the robot (all distance sensor values are equal to 0). The output produced in all cases for the motors is maximum forward speed (note the constant term of the approximator is 1 for both motor outputs $v_1$ and $v_2$). The value of the internal variable $y_1$ is set to 1 when light is present (rules 3 and 4) and to the previous value (can be 0 or 1) if no light is measured (rules 1 and 2). Rule 5 produces a turn to the right (left motor at maximum speed) if no light was detected before ($y_1 = 0$) and rule 6 a turn to the left (right motor at maximum speed) if light was detected ($y_1 = 1$). In both cases, the flag (internal variable $y_1$) is reset to 0. It is important to note that the apriori knowledge is defined by specifying rules that determine the expected behavior of the controller in specific points in the input domain, without specifying the area of application of the rules, as it was detailed in section~\ref{sec:properties}.

The results for the best and average fitness over 10 runs are listed in table~\ref{table:exp2}, where the best possible value for fitness is 1 and 0 is the worst.
\begin{table}
\centering
\caption{Fitness for the evolutionary robot experiment}
\begin{tabular}{|c|c|c|c|c|c|} \hline
\multicolumn{3}{|c|}{RFV} & \multicolumn{3}{c|}{RFV + apriori knowledge} \\ \hline
mean   & best   & var    & mean   & best   & var    \\ \hline
0.7728 & 0.8723 & 0.0686 & 0.8510 & 0.8835 & 0.0255 \\ \hline
\end{tabular}
\label{table:exp2}
\end{table}
Smaller error is achieved with apriori rules, but also the standard deviation is smaller, meaning that it is a more robust approach.

The figure~\ref{fig:evaluation} shows the performance of the best controllers found during evolution in a scenery not used during evolution. The controllers are evaluated for 800 time steps. The controller that do not use apriori knowledge can drive the robot for a longer distance in the same number of time steps, performing not so abrupt turns in the intersections but both controllers can drive the robots by following the light signals as expected.

\begin{figure}
\centering
\begin{tabular}{c}
\includegraphics[scale=0.6]{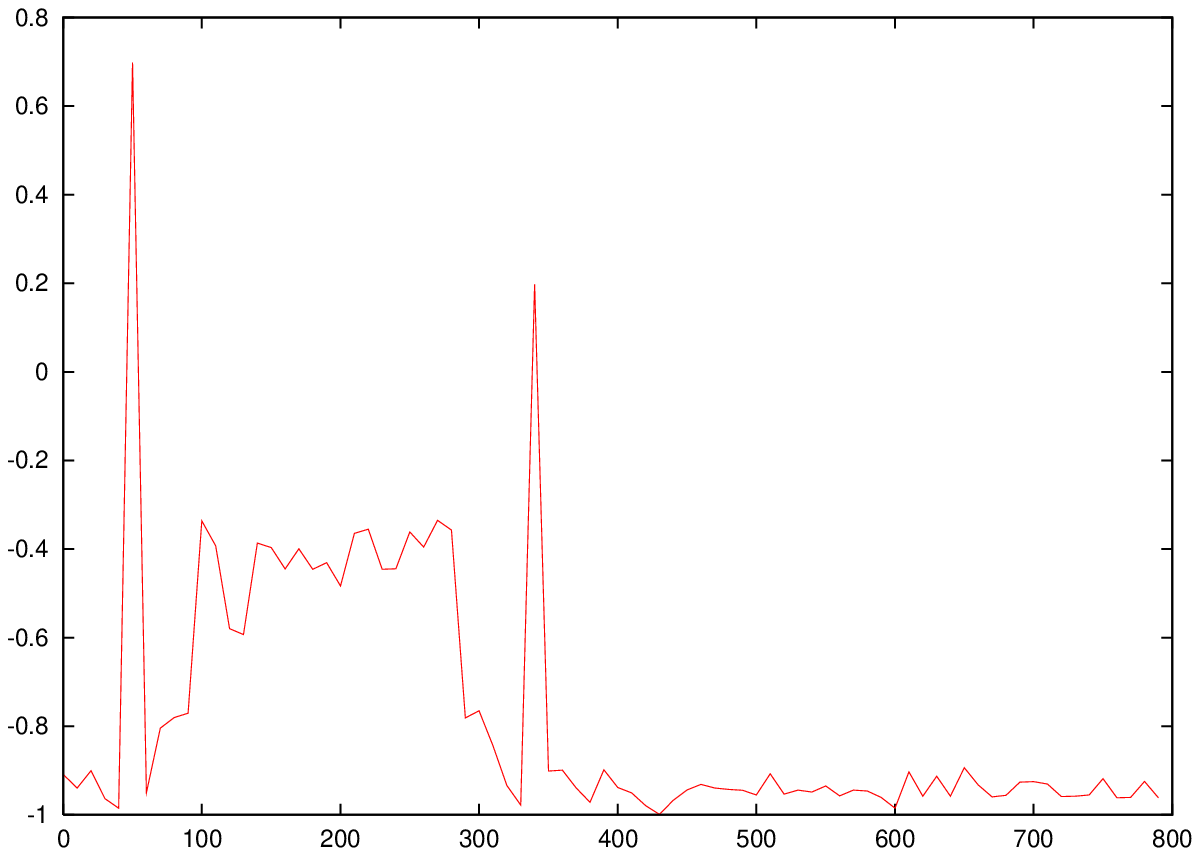} \\
(a) \\
\includegraphics[scale=0.6]{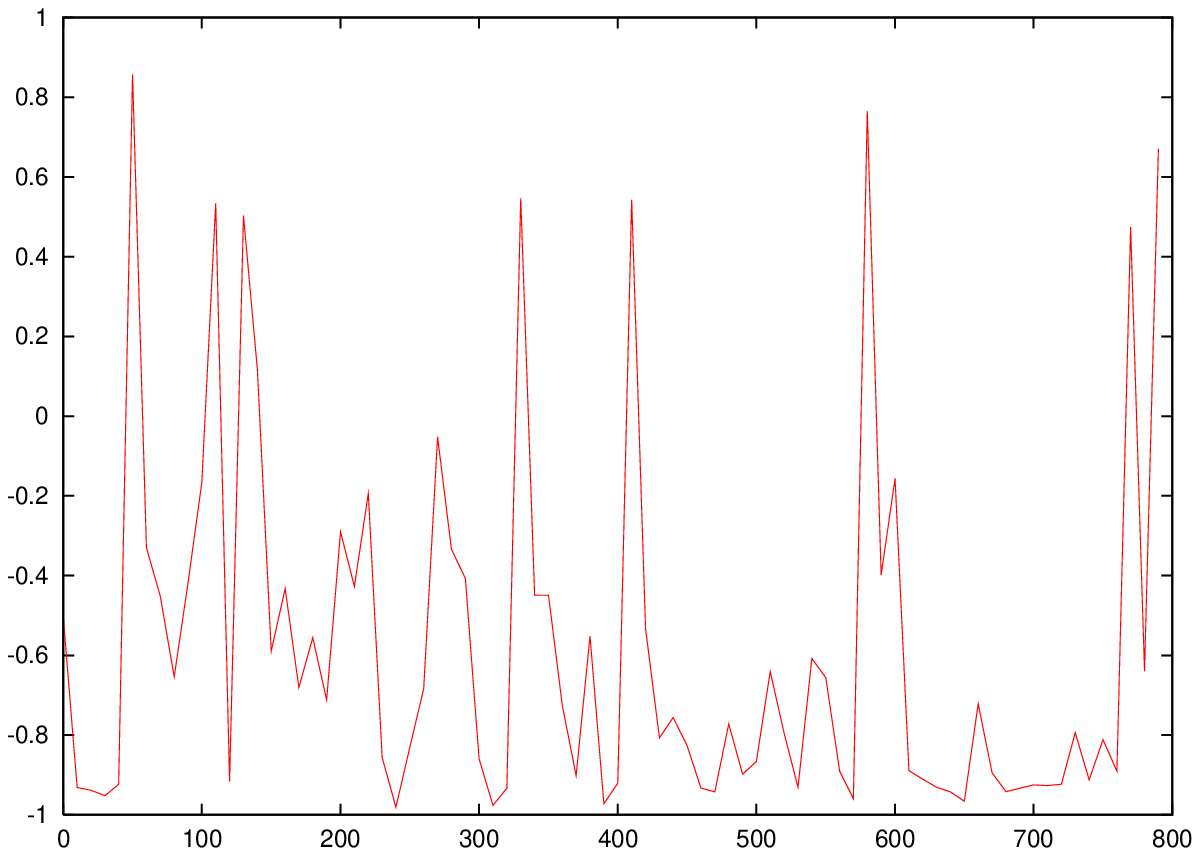} \\
(b) \\
\end{tabular}
\caption{Value of the hidden unit of the best controller obtained through evolution (a) without and (b) with apriori knowledge when evaluated on the test scenery.}
\label{fig:internal}
\end{figure}

However, the most important point is that a definite semantic interpretation of the hidden unit is provided with the apriori rules: the hidden unit behavior indicates if light was or not detected before the intersection. There is no guarantee that a clear semantic is provided with the approach without apriori knowledge. Figure~\ref{fig:internal} shows the value of the hidden unit of both best controllers plotted for the 800 time steps when evaluated on the test scenery from figure~\ref{fig:evaluation}.
The value of the hidden unit for the controller evolved with apriori knowledge represent the expected semantics, with two peaks on the areas where light signals were detected. No clear semantics can be defined in the case of the controller evolved without apriori knowledge.

\section{Conclusions}
\label{sec:conclusions}

In this paper, the RFV model has been proposed. This model is an extension of the FV model defined to extend the application domain to include temporal problems. The temporal relations are embedded by including internal units that provide feedback by connecting outputs to inputs. These internal units act as memory elements. This paper propose the use of genetic algorithms as a design tool of the RFV model. The controllers are represented by following the FV model, which is a representation for fuzzy controllers based on Voronoi diagrams that can represent fuzzy systems with synergistic rules, fulfilling the $\epsilon$-completeness property and providing a simple way to introduce apriori knowledge. The geometric interpretation of the rules allows the use of geometric genetic operators that proved to be useful also in other contexts. The main benefit of the proposed representation is the possibility to provide a definite semantic to the internal (or recurrent) units. The representation and the algorithms have been validated in two problems in the area of system identification and evolutionary robotics. Future work include experiments on a real mobile robot and the study of the impact of using the so-called {\em Symbolic Controllers} approach~\cite{godzik:evorobot}.

\bibliographystyle{abbrv}
\bibliography{paper-gecco-2005}

\begin{thebibliography}{10}

\bibitem{babuska:fuzzy}
R.~Babu\v{s}ka.
\newblock Fuzzy modeling: Principles, methods and applications.
\newblock In C.~Bonivento, C.~Fantuzzi, and R.~Rovatti, editors, {\em Fuzzy
  Logic Control: Advances in Methodology}, pages 187--220. World Scientific,
  Singapore, 1998.

\bibitem{carinena:temporal}
P.~Carinena, C.~Regueiro, A.~Otero, A.~Bugarin, and S.~Barro.
\newblock Landmark detection in mobile robotics using fuzzy temporal rules.
\newblock {\em IEEE Transactions on Fuzzy Systems}, 12(4):423--435, August
  2004.

\bibitem{berg:compgeometry}
M.~de~Berg, M.~van Kreveld, M.~OVermars, and O.~Schwarzkopf.
\newblock {\em Computational Geometry, Algorithms and Applications}.
\newblock Springer Verlag, 1998.

\bibitem{feng:adaptivecontrol}
G.~Feng.
\newblock An approach to adaptive control of fuzzy dynamic systems.
\newblock {\em IEEE Transactions on Fuzzy Systems}, 10(2):268--275, April 2002.

\bibitem{giles:equivalence}
C.~L. Giles, C.~W. Omlin, and K.~K. Thornber.
\newblock Equivalence in knwoledge representation: Automata, recurrent neural
  networks and dymical fuzzy systems.
\newblock {\em Proceedings of the IEEE}, 87(9):1623--1640, September 1999.

\bibitem{godzik:evorobot}
N.~Godzik, M.~Schoenauer, and M.~Sebag.
\newblock Evolving symbolic controllers.
\newblock In G.~R. et~al., editor, {\em Applications of Evolutionary
  Computing}, Lecture Notes in Computer Science 2611, pages 638--650, 2003.

\bibitem{hagras:type2}
H.~A. Hagras.
\newblock A hierarchical type-2 fuzzy logic control architecture for autonomous
  mobile robots.
\newblock {\em IEEE Transactions on Fuzzy Systems}, 12(4):524--539, August
  2004.

\bibitem{hojati:hybrid}
M.~Hojati and S.~Gazor.
\newblock Hybrid adaptive fuzzy identification and control of nonlinear
  systems.
\newblock {\em IEEE Transactions on Fuzzy Systems}, 10(2):211--221, April 2002.

\bibitem{juang:tsk}
C.-F. Juang.
\newblock A tsk-type recurrent fuzzy network for dynamic systems processing by
  neural network and genetic algorithms.
\newblock {\em IEEE Transactions on Fuzzy Systems}, 10(2):155--170, April 2002.

\bibitem{juang:inference}
C.-F. Juang and C.-T. Lin.
\newblock A recurrent self-organizing neural fuzzy inference network.
\newblock {\em IEEE Transactions on Neural Networks}, 10(4):828--845, July
  1999.

\bibitem{kasabov:denfis}
N.~Kasabov and Q.~Song.
\newblock Denfis: Dynamic evolving neural fuzzy inference systems and its
  application for time series prediction.
\newblock {\em IEEE Transactions on Fuzzy Systems}, 10(2):144--154, April 2002.

\bibitem{kavka:funcapp}
C.~Kavka and M.~Schoenauer.
\newblock Voronoi diagrams based function identification.
\newblock {\em Lecture Notes in Computer Science}, 2723:1089--1100, July 2003.

\bibitem{kavka:voronoi}
C.~Kavka and M.~Schoenauer.
\newblock Evolution of voronoi based fuzzy controllers.
\newblock {\em Lecture Notes in Computer Science}, 3242:541--550, September
  2004.

\bibitem{lee:control}
C.-H. Lee and C.-C. Teng.
\newblock Identification and control of dynamic systems using recurrent fuzzy
  neural networks.
\newblock {\em IEEE Transactions on Fuzzy Systems}, 8(4):349--366, August 2000.

\bibitem{lin:nnfs}
C.~T. Lin and S.~G. Lee.
\newblock {\em Neural Fuzzy Systems: A Neural-Fuzzy Synergism to Intelligent
  Systems}.
\newblock Prentice Hall, Englewood CLiffs, NJ, 1986.

\bibitem{mastorocostas:identification}
P.~A. Mastorocostas and J.~B. Theocharis.
\newblock A recurrent fuzzy-neural model for dynamic system identification.
\newblock {\em IEEE Transactions on Systems, Man and Cybernetics},
  32(2):176--190, April 2002.

\bibitem{michel:khepera}
O.~Michel.
\newblock Kephera simulator package version 2.0: Freeware mobile robot
  simulator.
\newblock http://wwwi3s.unice.fr/~om/khep-sim.html.

\bibitem{mouzouris:nfsls}
G.~Mouzouris and J.~Mendel.
\newblock Dynamic non-singleton fuzzy logic systems for nonlinear modeling.
\newblock {\em IEEE Transactions on Fuzzy Systems}, 5(2):199--208, May 1997.

\bibitem{nolfi:robotics}
S.~Nolfi and D.~Floreano.
\newblock {\em Evolutionary Robotics, The Biology, Intelligence, and Technology
  of Self-Organizing Machines}.
\newblock Bradford Books, 2000.

\bibitem{schoenauer:voronoi}
M.~Schoenauer, F.~Jouve, and L.~Kallel.
\newblock Identification of mechanical inclusions.
\newblock In D.~Dasgupta and Z.~Michalewicz, editors, {\em Evolutionary
  Algorithms in Engineering Applications}. Springer Verlag, 1997.

\bibitem{sun:robotics}
Y.~L. Sun and M.~J. Er.
\newblock Hybrid fuzzy control of robotic systems.
\newblock {\em IEEE Transactions on Fuzzy Systems}, 12(6):755--765, December
  2004.

\bibitem{zhang:nonlinear}
J.~Zhang and J.~Morris.
\newblock Recurrent neuro fuzzy models for nonlinear process modeling.
\newblock {\em IEEE Transactions on Neural Networks}, 10(2):313--326, March
  1999.

\end{thebibliography}

\end{document}